\documentclass[letterpaper]{article}
\usepackage[draft]{aaai2026}
\usepackage{times}
\usepackage{helvet}
\usepackage{courier}
\usepackage[hyphens]{url}
\usepackage{graphicx}
\usepackage{natbib}
\usepackage{caption}
\usepackage{algorithm}
\usepackage{algorithmic}
\usepackage{newfloat}
\usepackage{listings}
\usepackage{amsmath}
\usepackage{booktabs} 

\title{AI PB: A Grounded Generative Agent for Personalized Investment Insights}

\author{
    Daewoo Park, Suho Park, Inseok Hong, Hanwool Lee, Junkyu Park, Sangjun Lee, Jeongman An, Hyunbin Loh\textsuperscript{*}
}
\affiliations{
    Shinhan Securities AI Solution Dept, Seoul, South Korea\\
    \textsuperscript{*}Corresponding author: Hyunbin Loh 
}

\begin{document}
\maketitle

\begin{abstract}
We present \textit{AI PB}, a production-scale generative agent deployed in real retail finance. Unlike reactive chatbots that answer queries passively, AI PB proactively generates grounded, compliant, and user-specific investment insights. It integrates (i) a component-based orchestration layer that deterministically routes between internal and external LLMs based on data sensitivity, (ii) a hybrid retrieval pipeline using OpenSearch and the finance-domain embedding model, and (iii) a multi-stage recommendation mechanism combining rule heuristics, sequential behavioral modeling, and contextual bandits. Operating fully on-premises under Korean financial regulations, the system employs Docker~Swarm and vLLM across~24~X~NVIDIA~H100~GPUs. Through human QA and system metrics, we demonstrate that grounded generation with explicit routing and layered safety can deliver trustworthy AI insights in high-stakes finance.
\end{abstract}

\section{Introduction}
Generative AI promises to democratize financial intelligence, enabling anyone to query markets, portfolios, or company data through natural language. Yet, deploying LLM-based systems in finance introduces structural challenges: hallucination, unverifiable reasoning, and regulatory risk.\cite{son2023removingnonstationaryknowledgepretrained}  
Traditional chatbot deployments often rely on external APIs, making them incompatible with strict privacy regimes where personal trading or account information cannot leave the institution\cite{li2025rethinkingdataprotectiongenerative}.

\textbf{AI PB (AI Private Banker)} was conceived as a next-generation, enterprise-grade agent that combines the analytical depth of large models with the verifiability and control of classical systems.  
Its design objectives are:
\begin{enumerate}
\item \textbf{Ground-first generation:} all textual outputs are constructed from verified enterprise data.  
\item \textbf{Routing by component:} model choice is determined by predefined component specifications, ensuring deterministic compliance.  
\item \textbf{Proactive personalization:} daily insights are pre-generated from predicted user intents rather than waiting for queries.
\end{enumerate}

The system now serves as a live assistant in retail channels, offering users data-driven commentary, portfolio diagnostics, and market narratives in a conversational format.

\textbf{Contributions.}
\begin{itemize}
\item A deployed architecture comprising 20~components and 48~enterprise modules for financial reasoning and analytics.  
\item A routing policy guaranteeing zero PII egress by design.  
\item A hybrid retrieval mechanism that anchors every generation to factual evidence.  
\item A hybrid recommendation engine that personalizes insight delivery.  
\item Empirical evidence from human QA showing over 90\% factuality with near-perfect compliance.
\end{itemize}

\section{User Experience and Interaction Flow}
\textbf{Dual-interface design.}
AI PB provides two complementary user surfaces:
\begin{itemize}
\item \textit{Today Feed}—a dynamic dashboard that delivers pre-generated daily briefings, including disclosure digests, market movers, sector narratives, and personalized portfolio summaries.  
\item \textit{Dialogue View}—an interactive chat allowing free-form, multi-turn conversations with contextual grounding and visual elements.
\end{itemize}

The design allows seamless transition between passive consumption (feed) and active exploration (chat), creating an “always-on” advisor experience.

\begin{figure}[t]
\centering
\includegraphics[width=0.9\columnwidth]{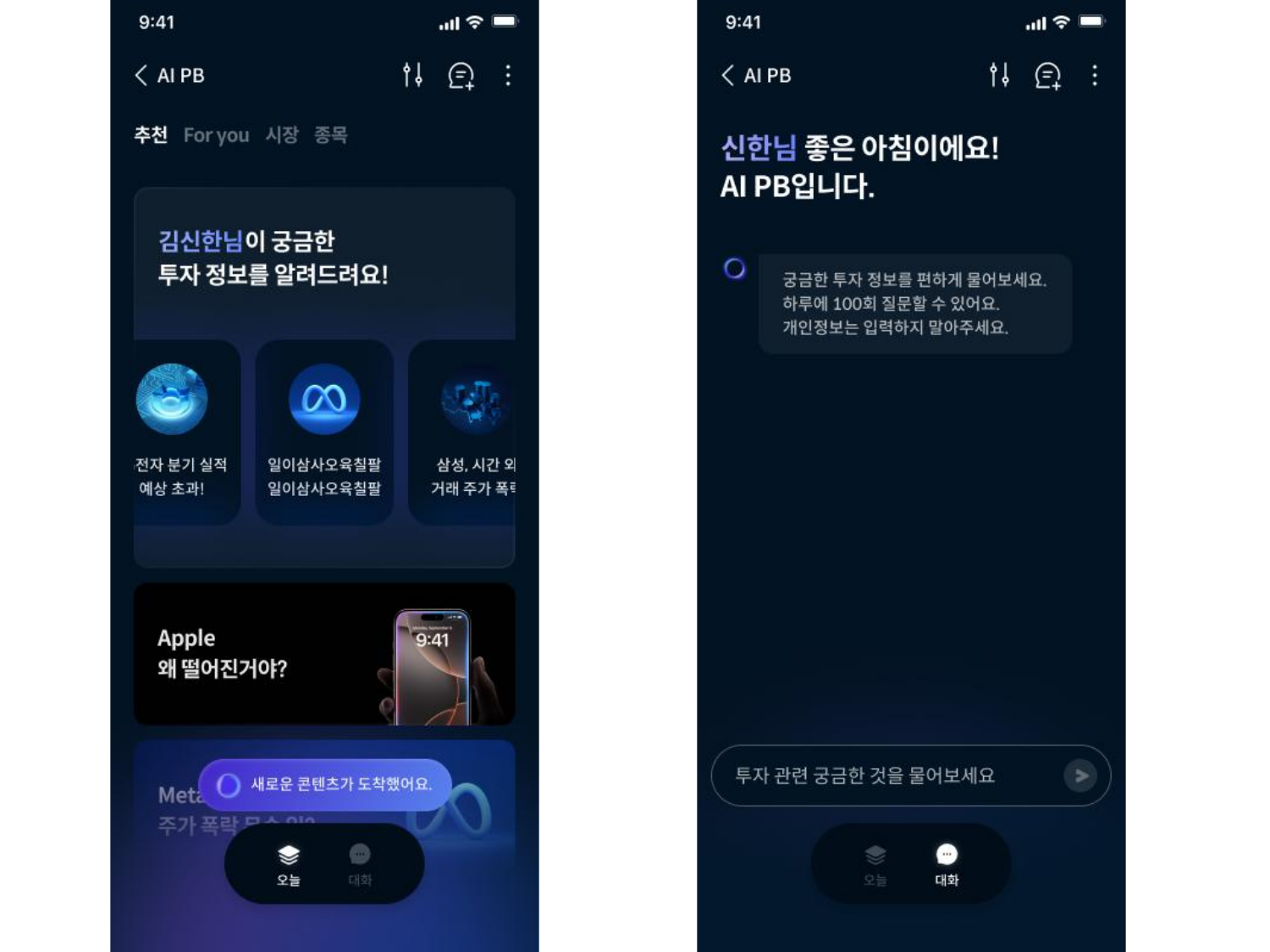}
\caption{\textbf{User interface of AI PB.} The ``Today'' view proactively surfaces grounded insights, while the ``Dialogue'' view enables multi-turn Q\&A and reasoning over financial data.}
\label{fig:ui}
\end{figure}

\section{System Architecture}
\label{sec:architecture}
\begin{figure}[t]
  \centering
  \includegraphics[width=0.9\columnwidth]{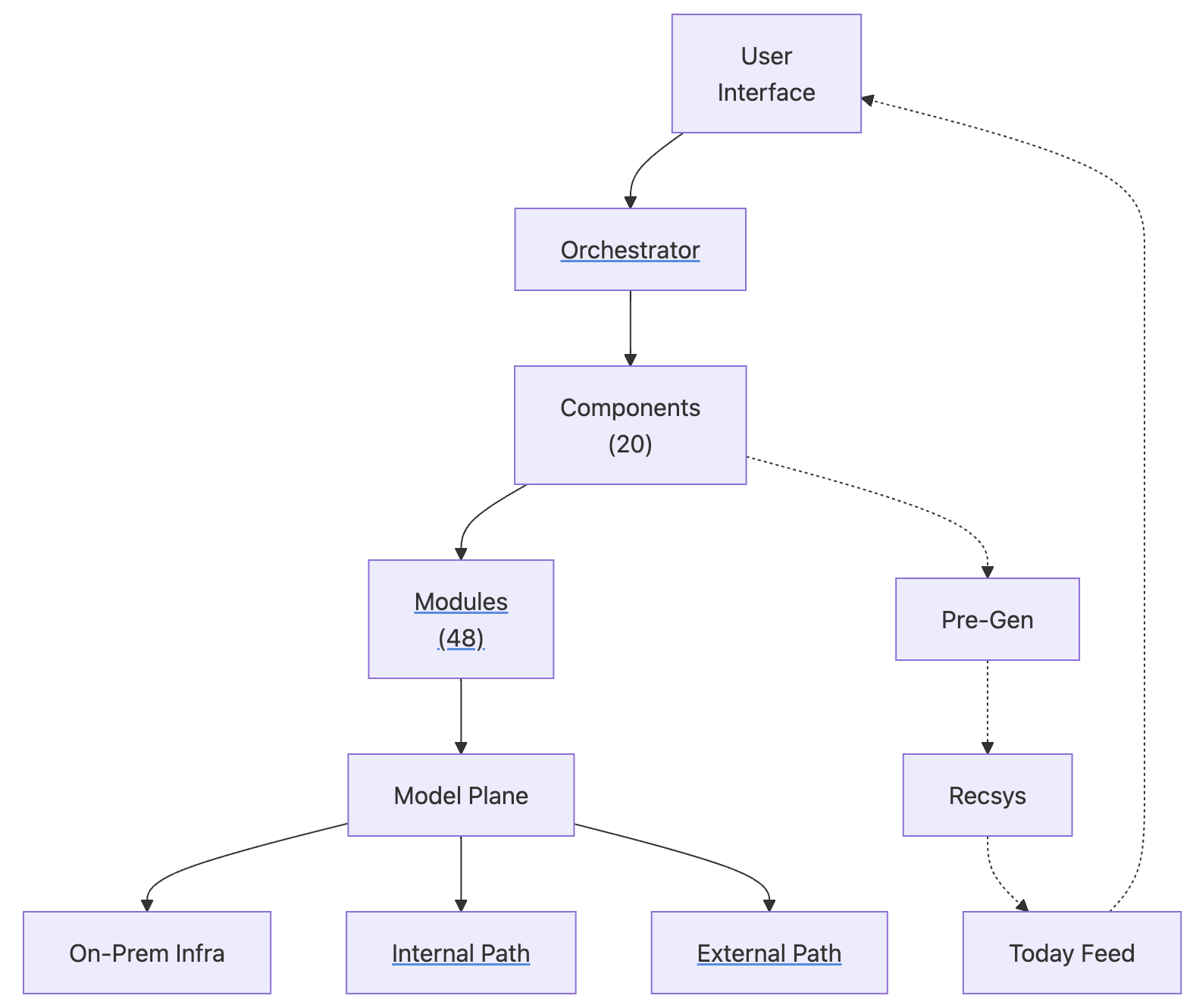} % .svg/.pdf/.png
  \caption{\textbf{System architecture of AI PB.} Requests flow from UI to Orchestrator, then Components and Modules. Components handling PII use the \emph{Internal} path; non-PII may use the \emph{External} path. Dashed side-flow shows Pre-Gen ranked into the \emph{Today Feed}.}
  \label{fig:architecture}
\end{figure}

AI PB’s backend consists of two cooperative subsystems:  
(1) the \textbf{Single-Agent System}, integrating 20~components (18~informational and~2~personal-asset analytic) on top of 48~data modules; and  
(2) the \textbf{Recommendation System}, which ranks and surfaces pre-generated insights (Figure~\ref{fig:architecture}).

\textbf{Component–Module hierarchy.}
In AI PB, a \textit{component} corresponds to a high-level \textbf{workflow} that captures a user intent (e.g., “analyze my portfolio” or “summarize recent disclosures”). 
Each component is composed of multiple \textit{modules}, which serve as executable \textbf{tools} interfacing with internal systems—such as data retrieval, analysis, summarization, and evidence packaging.  
This design allows logical workflows (components) to be flexibly assembled from reusable functional units (modules), improving maintainability and traceability. \cite{park2025practicalapproachbuildingproductiongrade}

\textbf{Component workflows.}
Each component represents a logical workflow—\textit{Company Overview}, \textit{Disclosure Summary}, \textit{Financial Ratios}, \textit{Theme Leaders}, or \textit{Portfolio~Analysis}.  
Modules access data from AltiBaseDB, OracleDB, PostgreSQL, OpenSearch, and legacy TR systems.  
The architecture is modular enough to allow independent scaling and versioning of each component.

\section{Component-Level Orchestration and Safety}
Routing is handled by a deterministic policy rather than runtime prompt inspection.  
When a user query arrives:
\begin{enumerate}
\item The orchestrator identifies the invoked component and checks its metadata on data sensitivity.  
\item If the component accesses any PII-linked source, generation occurs via the \textbf{internal model path}.  
\item Otherwise, the system may call an \textbf{external model path} (e.g., GPT-4o) for stylistic generation.
\end{enumerate}

\textbf{Safety Guard.}
All inputs and outputs are filtered through \textit{Shinhan-Guard}, a fine-tuned derivative of Llama~Guard~3~\cite{dubey2024llama3herdmodels} adapted to Korean financial contexts.  
Rejections trigger fallback to safe templates, such as compliance disclaimers or generalized summaries.  
This guard-based rejection mechanism ensures zero policy-violating text reaches end users.

\textbf{Guard performance.}
Shinhan-Guard was evaluated across multiple public and in-house safety benchmarks to assess its robustness in regulated financial dialogue.  
All metrics are reported as \textbf{F1~scores}:

\begin{itemize}
\item \textbf{Toxicity (in-house benchmark):} 0.8233 (F1)  
\item \textbf{HarmBench (standard)}~\cite{mazeika2024harmbenchstandardizedevaluationframework}: 0.985 (F1)  
\item \textbf{HarmBench (contextualized)}~\cite{mazeika2024harmbenchstandardizedevaluationframework}: 0.94 (F1)  
\item \textbf{Safe-Guard Prompt Injection}~\cite{xTRam1_safe_guard_prompt_injection_2025}: 0.90 (F1)  
\item \textbf{PII Detection Benchmark} (in-house benchmark): 0.9488 (F1)  
\end{itemize}

These results indicate that Shinhan-Guard maintains high recall on harmful or privacy-sensitive content while preserving fluent responses, achieving near-perfect robustness on HarmBench and strong generalization to contextual and in-domain safety risks.

\section{Grounded Retrieval and Evidence Integration}
Every answer produced by AI PB must be verifiable against enterprise data.  
The system therefore employs a \textbf{hybrid retrieval pipeline} combining\cite{mala2025hybridretrievalhallucinationmitigation}:
\begin{itemize}
\item OpenSearch-based sparse retrieval for symbolic keyword matching,  
\item NMIXX\cite{lee2025nmixxdomainadaptedneuralembeddings} dense retrieval for semantic relevance\cite{hwang2025twiceadvantageslowresourcedomainspecific}, and  
\item Query expansion using domain ontologies and temporal normalization.
\end{itemize}

Retrieved passages are serialized into evidence templates and provided to the generator as structured context.  
A post-generation validator ensures that each statement contains at least one reference token.  
This approach reduces hallucination by over~30\% relative to vanilla prompting in internal tests.

\section{Proactive Personalization and Recommendation}
AI PB continuously generates 22~types of daily insights per user through predefined templates such as ``top contributors to portfolio return'' or ``disclosure alerts for watched companies.''  
The resulting pool is ranked by a three-layer hybrid recommender designed to balance relevance, diversity, and novelty.

\textbf{Rule-based layer.}  
This layer prioritizes owned and watched tickers while down-weighting items the user has already read.  
It also applies time-decay rules to keep content fresh and prevent repetitive exposure.  
These deterministic rules enforce domain logic and ensure that essential disclosures always appear.

\textbf{Sequential RS.}  
The sequential model predicts the next likely interest based on a user’s recent reading sequence and interaction patterns.\cite{sun2019bert4recsequentialrecommendationbidirectional}  
It captures temporal intent shifts---for example, moving from sector summaries to company-level diagnostics---and refines personalization over time.

\textbf{Contextual bandit layer.}  
Finally, a contextual multi-armed bandit adjusts rankings in real time using click and dwell feedback.\cite{Li_2010}  
It operates under a limited trust budget, allowing controlled exploration while maintaining consistency and reliability.

In A/B testing, this hybrid approach increased daily feed engagement by~18\% and reduced repetitive content by~23\% compared with rule-based baselines.

\section{Deployment and Serving}
All components operate within an on-premises data center to ensure full data sovereignty.  
The system runs on a Docker~Swarm cluster using \textit{vLLM}\cite{kwon2023efficientmemorymanagementlarge} for efficient inference across~24~NVIDIA~H100~GPUs.  
Eight GPUs serve guard, embedding, and reranking tasks, while sixteen handle generative workloads based on the \textbf{Qwen3-32B~ORPO-aligned} model.

\textbf{Model alignment.}
To maintain stylistic and factual consistency between external and internal models, we performed \textbf{LoRA+ORPO}\cite{hu2021loralowrankadaptationlarge, hong2024orpomonolithicpreferenceoptimization} tuning.  
Roughly~10{,}000~synthetic gold outputs were generated and manually reviewed by domain experts to reflect financial safety and tone requirements.  
This dataset was used to optimize both regulatory compliance and user alignment without degrading fluency.

\textbf{Embedding and retrieval.}
The retrieval layer uses the \textbf{NMIXX} embedding model\cite{lee2025nmixxdomainadaptedneuralembeddings}, specialized for financial semantics.  
NMIXX captures contextual meaning shifts—such as temporal valuation or regulatory framing—allowing more accurate hybrid retrieval from enterprise OpenSearch indices.

\textbf{Serving pipeline.}
vLLM enables continuous batching and KV-cache reuse for conversational inference, minimizing latency even under high concurrency.\cite{kwon2023efficientmemorymanagementlarge}  
Each microservice container scales independently through Swarm’s overlay network.  
Logs, routing traces, and guard verdicts are collected in a Prometheus–Grafana stack for live monitoring.

\textbf{Performance.}
Pre-generation runs asynchronously, while interactive chats execute synchronously with caching and grounding verification.  
Indices update hourly to reflect new disclosures.  
The full system sustains tens of thousands of daily events with p95 latency under~5.9~seconds and a guard rejection rate below~2\%.

\begin{table}[t]
\centering
\caption{Representative serving profile (anonymized).}
\label{tab:serving}
\begin{tabular}{l r}
\hline
Category & Value \\
\hline
Guard / Embedding / Reranker GPUs & 8 × H100 \\
Generation GPUs (Qwen3-32B ORPO) & 16 × H100 \\
Average Pre-gen Latency & 5.9~s / item \\
Average Chat Latency (p95) & 13.9~s \\
Guard Rejection Rate & 1.8\% \\
Retrieval Refresh Interval & $\le$15~min \\
\hline
\end{tabular}
\end{table}

\section{Evaluation}

\textbf{Component Classification Metric.}
To verify whether the orchestrator correctly invokes the intended component, we designed a custom metric evaluating both precision and coverage of component selection.  
Given a user query $q$, the model outputs a set of predicted components $C_p$, while the human-annotated gold components are $C_g$.  
We define:
\[
\text{Score} = 
\alpha \left( \frac{|C_p \cap C_g|}{|C_g|} \right) +
\beta \left( 1 - \frac{|C_p \setminus C_g|}{|C_p|} \right),
\]
where $\alpha$ and $\beta$ balance correctness and over-generation penalties.
The first term rewards accurate component hits, while the second penalizes irrelevant ones.  
For questions with multiple valid components, this formulation captures both completeness and precision of routing.

Evaluation was conducted on approximately 5{,}000~user queries collected from public investor forums.  
Each query was manually labeled with its gold component list among 20~predefined components.  
The model achieved an average routing score of \textbf{510.24}(When $\alpha = 0.5$, $\beta = 0.5$), demonstrating stable and reliable component selection within the orchestration framework.

\textbf{Human QA.}
Two professional QA reviewers evaluated over 300~randomly sampled responses across components under a three-axis rubric:
\begin{enumerate}
\item \textbf{Factuality:} correctness and citation alignment.  
\item \textbf{Safety:} absence of policy or privacy violations.  
\item \textbf{Alignment:} linguistic clarity and user helpfulness.
\end{enumerate}
Inter-rater agreement ($\kappa=0.78$) confirmed reliability.  
Average factuality reached~91.2\%, safety~98.4\%, and alignment~85.7\%.

\textbf{System metrics.}
Automatic logging tracks guard triggers, latency percentiles, and citation coverage.  
Groundedness is quantified as the proportion of sentences linked to retrievable sources.  
Latency traces show minimal variance despite hybrid retrieval overhead.

\section{Discussion and Lessons Learned}
\textbf{Grounding vs responsiveness.}
Extensive retrieval improves factuality but increases latency; proactive pre-generation amortizes this cost by preparing contextual summaries ahead of demand.

\textbf{Auditability through routing.}
Component-level routing creates transparent audit trails.  
Each invocation logs the component ID, model used, and guard verdict, enabling full traceability for regulators.

\textbf{Human-in-the-loop.}
Regulatory contexts evolve faster than policy models.  
Maintaining a QA loop ensures adaptation and provides qualitative feedback beyond metrics.

\section{Conclusion}
AI PB demonstrates that grounded, compliant, and personalized generative agents can be deployed at enterprise scale.  
By combining deterministic routing, hybrid retrieval, and layered safety, it achieves both factual reliability and user trust.  
We hope this case study offers a blueprint for deploying LLM-based agents in other regulated industries such as banking and healthcare.

\section*{Ethical Statement}
All experiments comply with institutional privacy and communication policies.  
External models never receive personally identifiable information.  
All outputs are informational and non-advisory; the system does not provide investment recommendations.

\appendix
\section*{Contributors}
\vspace{-1mm}
The development of \textit{AI PB} was a collaborative effort across multiple teams within Shinhan Securities. 
The following individuals contributed to the design, engineering, and deployment of the system.

\begin{center}
\renewcommand{\arraystretch}{1.1}
\setlength{\tabcolsep}{8pt}
\begin{tabular}{p{0.32\linewidth} p{0.6\linewidth}}
\toprule
\textbf{Role} & \textbf{Name(s)} \\
\midrule
Product Owner & Daewoo Park \\
Product Manager / Service Planning & Suho Park \\
Service Planning & Jihyun Lee \\
Backend Co-Leads & Inseok Hong, Junkyu Park \\
Backend Engineer & Kiwoong Ko, Junwoo Lim \\
AI Lead & Hanwool Lee \\
AI Engineers & Minsu Jung, Jaehun Cho \\
Frontend Lead & Sangjun Lee \\
Quant Lead & Eunho Choi \\
Quant Engineer & Nakyoung Lee \\
Data Engineering Lead & Jeongman An \\
Data Engineer & Chanhoe Lee \\
UI & Yeonju Park \\
QA & Seohyun Park, Minju Kim \\
\bottomrule
\end{tabular}
\end{center}

\vspace{-2mm}
\smallskip
\noindent\textit{Note.} Listed in functional order. All contributors participated in the deployment and testing of the production-scale system.

\bibliography{aaai2026}
\end{document}